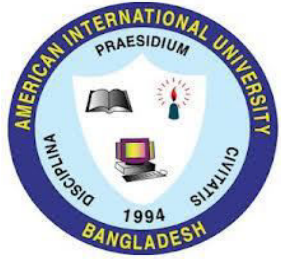

# Automated Toll Management System Using RFID and Image Processing

**Project Thesis**

**Submitted By**

| 19-40978-2 | AHMED, RAIHAN |
| --- | --- |
| 18-38760-3 | OMI, SHAHED CHOWDHURY |
| 18-38832-3 | RAHMAN, MD. SADMAN |
| 18-38770-3 | BHUIYAN, NIAZ RAHMAN |

**Department of Computer Science**

**Faculty of Science & Technology**

**American International University Bangladesh**

**December, 2022**

# Declaration

We declare that this thesis is our original work and has not been submitted in any form for another degree or diploma at any university or other institute of tertiary education. Information derived from the published and unpublished work of others has been acknowledged in the text and a list of references is given.

\_\_\_\_\_\_\_\_\_\_\_\_\_\_\_\_\_\_\_\_\_\_\_\_\_\_\_\_\_\_  \_\_\_\_\_\_\_\_\_\_\_\_\_\_\_\_\_\_\_\_\_\_\_\_\_\_\_\_\_\_

**AHMED, RAIHAN**  **OMI, SHAHED CHOWDHURY**
19-40978-2  18-38760-3
Computer Science and Engineering  Computer Science and Engineering
Faculty of Science and Technology  Faculty of Science and Technology

\_\_\_\_\_\_\_\_\_\_\_\_\_\_\_\_\_\_\_\_\_\_\_\_\_\_\_\_\_\_  \_\_\_\_\_\_\_\_\_\_\_\_\_\_\_\_\_\_\_\_\_\_\_\_\_\_\_\_\_\_

**RAHMAN, MD. SADMAN**  **BHUIYAN, NIAZ RAHMAN**
18-38832-3  18-38770-3
Computer Science and Engineering  Computer Science and Engineering
Faculty of Science and Technology  Faculty of Science and Technology

# Approval

This thesis titled "Automated Toll Management System Using RFID and Image Processing" has been submitted to the following respected member of the board of examiners of the Faculty of Science and Technology impartial fulfillment of the requirements for the degree of Bachelor of Science in Computer Science and Engineering on dated and has been accepted satisfactory.

\_\_\_\_\_\_\_\_\_\_\_\_\_\_\_\_\_\_\_\_\_\_\_\_\_\_\_\_\_\_       \_\_\_\_\_\_\_\_\_\_\_\_\_\_\_\_\_\_\_\_\_\_\_\_\_\_\_\_\_\_
**RASHIDUL HASAN NABIL**                **MD. KISHOR MOROL**

Lecturer & Supervisor                   Assistant Professor & External
Department of Computer Science          Department of Computer Science
American International University-Bangladesh   American International University-Bangladesh

\_\_\_\_\_\_\_\_\_\_\_\_\_\_\_\_\_\_\_\_\_\_\_\_\_\_\_\_\_\_       \_\_\_\_\_\_\_\_\_\_\_\_\_\_\_\_\_\_\_\_\_\_\_\_\_\_\_\_\_\_
**DR. MD. ABDULLAH-AL-JUBAIR**          **DR. DIP NANDI**

Assistant Professor & Head (Undergraduate)   Professor & Director
Department of Computer Science          Faculty of Science & Information Technology
American International University-Bangladesh   American International University-Bangladesh

\_\_\_\_\_\_\_\_\_\_\_\_\_\_\_\_\_\_\_\_\_\_\_\_\_\_\_\_\_\_
**MASHIOUR RAHMAN**

Sr. Associate Professor & Associate Dean
Faculty of Science and Technology
American International University-Bangladesh

# TABLE OF CONTENTS

# LIST OF TABLES & FIGURES

# Abstract

Traveling through toll plazas is one of the primary causes of congestion, according to a recent study that examined these factors [29]. Electronic Toll Collection (ETC) can be set up to reduce the problem. This experiment focuses on improving the security of ETC by RFID tag, and number plate checking. For number plate checking image processing is used where CNN Classifier was implemented in order to detect vehicle registration numbers from the number plate. according to the registered number plate, mail will be sent to the respective owner for paying a toll fee within a certain amount of time to avoid a potential fine. While in this process toll fee will be automatically deducted in real-time from the owner's balance. People who would be waiting in queues to pay tolls would ultimately benefit from this system because no vehicles would slow down, saving them important time and inconvenience while traveling.

# 1. Introduction

Traveling via toll plaza becomes a hassle for travelers. A recent study has analyzed current reasons for congestion. The delay and queue are mainly due to two different charging methods namely known as "Manual Toll Collection (MTC)" and "Electronic Toll Collection (ETC)". In an Electronic Toll Collection, the user will be able to pay the fee using their credit card which in turn becomes time-consuming. All the lanes are having mixed lane system which is both MTC and ETC vehicles pass through the same lane. Tollbooth capacity and type of toll service have an influence on traffic operations and the overall efficiency of the toll plaza. During this process vehicles are required to halt thus making a queue to pay the toll fee which causes time delay and may also be generating a traffic jam. Traffic jams are a huge problem around the globe for which people are not being able to reach their destinations on time. Heavy vehicles transport commodities from the state, district, or province to their desired destinations. In most of the scenarios, there remains a common scenario where a huge line is piled up for people paying tolls. Mins or even hours are spent at a time on the same spot in order to pay the toll amount and move on. Mostly the vehicles that travel with commodities like groceries, fish, vegetables, or raw materials are being affected. This damage to items in turn results in significant losses across multiple businesses. Furthermore, toll plazas are also requiring continuous manpower to operate in order to collect tolls manually. It eventually causes significant delays and human errors. The project which has been designed is to develop an automated system that will check if a passing vehicle's RFID (Radio-Frequency Identification) tag is activated or disabled. If the chip remains initiated then will match passenger information from the database. While in this process toll fee will be automatically deducted in real-time from the owner's balance. And if the tag is inactive or not found then the laser will trigger the camera to capture two pictures of the number plate from the front and rear. The information collected from images of the number plate; a mail will be forwarded to the owner for clearing fees in order to avoid any fines. People who would be waiting in queues to pay tolls would ultimately benefit from this system because no vehicles would slow down, saving them important time and inconvenience while traveling.

The common problem with the current manual toll management system as well as the already existing automated toll management system can be solved by using an automated online payment system that will allow all the vehicles to pass the toll plaza even if the RFID tag in the vehicle is not available. This doesn't mean that the unregistered vehicles don't have to pay. In this system whenever any vehicle approaches the toll plaza, the laser detector detects the vehicle. Then it will check if the vehicle's RFID (Radio-frequency identification) tag is activated or not. If the chip is activated then it will check the user's information from the database and deduct the balance from the car owner's account automatically. When the tag is inactive or the tag is not found the laser will trigger the camera and the camera will take two photos, one on the rear number plate and another on the license plate. Then according to the registered number plate, mail will be sent to the respective owner for paying the money within a certain amount of time to avoid a fine. Using this RFID tag, both time consumption is decreased and extra security measures are ensured. In this way, people don't have to stop at the toll gate even if they don't have registered for an RFID tag. Authority can still collect the fee from the unregistered users. This will completely remove any kind of traffic congestion at the toll gate. If a registered vehicle is stolen, the user can report the theft, and all toll plazas will be notified if that vehicle goes through. If the car goes through a toll plaza, local authorities will be notified of the vehicle's last known location, allowing them to monitor and locate it. There is software that can do these tasks only if the vehicle is registered and it has an RFID tag on the windshield. If the vehicle is unregistered then the owner has to manually pay the toll fee. But the difference between our system and the existing system is that our system will detect the vehicle owner even if the vehicle is unregistered to our system and send a mail to the respective owner informing him to pay the fee in time to avoid fines.

# 2. Literature Review

## 2.1 Benefits of Automated Toll Management

When opposed to the conventional manual approach, the electronic toll-collecting system (ETC) offers a number of advantages. The physical toll collection system requires the collector to accept the toll at a booth pausing each vehicle, which results in a stoppage in time, significant traffic blockage, and a long line of traffic. An electronic toll-collecting system does away with all of these problems. The introduction of a contemporary automated toll-collecting system in Bangladesh can therefore help solve the primary issue with the human toll-collection method [32]. Using automated toll collection gates will help improve the toll service by saving time, fuel and gas emissions, and corruption [11]. The bulk of the key informant stated that the manual toll collecting system's cash conversion process takes a lot of time. When cash change is unavailable, it might lead to unwarranted altercations between the driver and the toll collector. Some powerful individuals desire to avoid paying the fee. Some cars operated by particular government agencies are free from toll payments. But some of the automobiles owned by other government agencies also benefit from toll exemption. Sometimes the cash register malfunctions, which makes it difficult to collect tolls. The payment is delayed as a result of this issue. One of the main causes of the terrible traffic bottleneck in the toll booth is this. The toll plaza is quite time-consuming due to a huge line of automobiles. They firmly believe that ETC would aid in reducing traffic bottlenecks caused by lengthy lines for manually paying tolls [32]. It also has the advantage that the government does not lose revenue from toll collection. In addition, the other option of bridge weight sensor always helps to avoid all kinds of undesirable casualties such as bridge collapse [29]. Benefits to the automated toll system include decreased operating costs, reduced labor costs, decreased maintenance costs, and improved cash management. The user benefits include time savings because there is no longer a need to search for change or accelerate or decelerate because the vehicles are not stopped to pay the toll. Additionally, there is a time saving as a result of the shorter toll transaction and waiting times. Furthermore, the removal of acceleration results in significant fuel savings. The decrease in mobile emissions that affect the areas close to where ETC is installed is one of the social advantages [45]. Another benefit that this automated toll collection will bring to increase in capacity increasing by three folds, accommodating traffic without requiring holding additional lanes.

## 2.2 License plate detection

Prior to detecting the license plate, the vehicle is detected from the video and its images are extracted. From these images license plates are detected using the template matching technique [31]. Many image segmentation algorithms are out there based on deep learning models, which have shown good performance on various image segmentation tasks and benchmarks. These algorithms are classified according to architecture types. There are various systems used to extract license plates dividing into different modules like pre-processing, plate detection, segmentation of the LP characters and many more. If there are multiple vehicles inside the frame at the same time it can detect all of the license plates and process them [1]. As stated in the paper, many problems arise during license plate detection, such as vehicle movement, noisy background, distance change, etc; Therefore, edge analysis method is used [5]. The invention relates to a system suitable for embedded systems for Brazilian license plate detection and recognition using a complex neural network (CNN). The resulting system detects license plates in images captured with the Tiny YOLOv3 architecture and characterizes them using a second complex network trained on composite images and fine-tuned with Real license plate image 'registered. To avoid the need for millions of license plate images, the solution of choice is to create a composite image of license plate images in 8-bit grayscale format. Advances in synthetic license plate images have reduced system development costs and still allow CNNs to achieve high accuracy [6]. Utilizing the data set to identify license plates in the input image is the initial phase of another system. Convolutional neural network (CNN) modeling was employed in this process. For a second, related CNN model to segment and identify license plate numbers, the collected images of license plates were used as fresh inputs. On 200 photos of him, they tested the model, and they got an accuracy of 99.5. In order to extract license plates for recognition and recognize license plates, they employed a CNN-based approach [7]. Two fully foldable single-stage object detectors were suggested in one research [36] to concurrently detect and identify signs and license plates using ALPRNet. As a result, less Optical Character Recognition (OCR) software use the Recurrent Neural Network (RNN) branch (OCR). This framework is designed to recognize mixed-style license plates. Applying ALPRNet achieves an accuracy of 98.21%. The OKM-CNN model is an effective model. It uses deep learning to recognize vehicle license plates. Another method can be recording continuous video of the oncoming traffic to take the best frame of the license plate. This technique is used in a system where Exemplar-SVM and convolutional neural networks with region proposal (RCNN) are applied in this paper [39] to detect license plates by treating them as an object. This system can detect license plates from the video even when the license plate is not fully visible. Three main steps of an OKM-CNN model were examined in a paper: license plate detection, clustering-based segmentation, and CNN-based license plate number recognition [8]. The image is preprocessed, then transformed to a binary image, where the characters from the license plate are segmented and extracted using the bounding box technique and the contours are recognized and filtered. then the characters are recognized using CNN [15]. The toll booth will be open round-the-clock. Therefore, the system ought to be able to process photos taken under dim lighting. Fog, rain, and storms shouldn't have an impact on how well the image processing performs [16]. As the Bangladeshi license plates only consist of Bengali letters and punctuation marks, an automatic license plate recognition system has been developed primarily for Bangladesh in which the authors have acquired colored images using a high-resolution camera, then they have pre-processed the image using various techniques and subsequent equations. Bangla license plate contains a hyphen, so they used connected components to segment the binary image. After segmentation, they used CNN to identify the characters in the number plate, their computer network infrastructure contains three convolutional layers: three maximum aggregation layers, three Rectified Linear Unit layers, fully connected layers and one classifier layer. Dataset for Bangla digit characters, to train and test their model [18]. As the vehicles do not need to stop or slow down at the toll plazas, the captured images can be blurry and unclear. So, to get better visibility super-resolution technique is used. Afterward bounding box method can be applied to partition each character from the image [31].

## 2.3 Automated Toll Collection

### 2.3.1 RFID Technology

Radio Frequency Identification (RFID) is a form wireless communication that lets users identify objects or people which are tagged uniquely with the use of transceiver and a transponder [42]. It can be used in various sectors such as agriculture, the manufacturing industry, keeping track of vehicles, and many more [43]. This technology consists of a two-part tag and a reader, the RFID reader uses a radio signal to read and get vehicle details saved on the RFID tag which is attached to the vehicle's windshield. All the items used in this system is very compact and easy to source. The price of these components is also very cheap. A memory chip is the core component of the RFID tag. This chip is used as a transceiver which is used to send a signal to activate the RFID tag and receive the information after decoding [17]. The range of an RFID reader is between (10 cm-30m) which is also important for the RFID tag to be detected effectively from an ideal range [20]. Thus, it can be stated that to detect vehicles RFID receivers should be kept at an optimum distance where RFID readers could read them easily.

### 2.3.2 Arduino

The Arduino is a microcontroller board that contains 16 analog inputs, serial ports, 54 digital I/O pins, a USB connector [17]. It also has a 16 MHz oscillator. Arduino is very useful for any specific work which is required to be done efficiently. It has a low cost in production and distribution as a result, the systems can be implemented at a cheaper rate [11]. Memory in static RAM is unstable. That is, the information stored in SRAM memory is lost if the microcontroller loses power. During the course of a program, it can both be written to and read from. 2K bytes of SRAM are available on the ATmega328. In addition to the input/output on the microcontroller, S-RAM is reserved for the registers utilized by the CPU. There is a comprehensive register listing for the ATmega328, along with a header file, in. RAM is used to house global variables, provide memory allocation and provide a space for the stack while a program is running [43].

A precise time basis called the clock regulates how quickly a microcontroller cycles through these processes. All peripheral subsystems receive a time basis from the microcontroller by way of a clock source that is distributed throughout the device. The ATmega328 microcontroller is either internally or externally timed using a Resistor Capacitor (RC) time basis. Programmable fuse bits are used to choose the internal time base of the RC. The internal fixed clock operating frequency can be set to 1, 2, 4, or 8 MHz. An external time source might be utilized to offer a larger variety of frequency options. The time base frequency and clock source device are selected by the system designer based on their suitability for the given application [41]. The RFID scanner which will be used in the automated toll collection system can be controlled by Arduino, as it is a low-cost solution for this project.

### 2.3.3 Laser Sensor

An electronic device called a laser sensor uses a concentrated light beam to determine if an object is present, absent, or far away. The physical measured value is converted into an analog electrical signal by a laser sensor, a measurement value recorder that uses laser technology. In other words, the laser sensor was designed for noninvasive measurement. The triangular principle incorporates how the laser sensor operates. Primarily laser light contains light waves of similar wavelengths which results in beam directing in a parallel direction. This light transmits information across long distances in order to detect objects or accurate distances. Nevertheless, in our project when a vehicle will be passing the laser transmitter a ray of red light would be penetrated which is eventually connected to a receiver.

### 2.3.4 Dataset Collection

Data collected from Kaggle contains many vehicle license plate images. There are two folders in the dataset one is called images which includes 433 pictures of car license plates and the other is called annotations which contains 433 xml files for each image's bounding box annotation. Images are taken from various angles which can be used to train the model, this training will be helpful during the extraction of the license plate image even when the angle of the car is not aligned perfectly during the image capture.

**2.3.5 Toll Management**

Using RFID sensors, Arduino Mega board and some cameras an automated toll system can be developed which can collect toll on the go. It will not require the vehicle to stop in the booth and even no toll booth will exist creating a seamless experience for the drivers. Image dataset can be used to train a deep learning model to detect vehicle from camera footage and classify number plate from the image and extract the license plate number [31]. Also, using RFID tag vehicles registration number can be found providing a reliable mechanism to detect vehicle's owner and auto deduct the toll charge from the owners account informing about the transection through SMS. All these information and transections can be stored in a cloud database and accessed using an android or iOS app and even a dedicated web app [35]. There are other techniques as well to implement an automated toll system using GPS and GPRS [19], ZigBee [23] and also using QR Code [28].

# 3. Dataset

A popular Car License Plate Detection dataset from Kaggle is applied to train the pre-trained object detection model of Tensorflow Object Detection API from ModelZoo to detect license plate of a vehicle. The dataset contains 433 images with bounding box annotations of the car license plates within the image. In xml files, the annotations are provided in the PASCAL VOC format. The dataset is then split into two parts, train and test. Where, train contains 411 images to train the model and test contains 22 images to evaluate the model [46].

# 4. Methodology:

The automated toll management system is a digital method of collecting tolls from vehicles passing through toll plazas. This method reduces corruption and the length of the vehicle queue.

## 4.1 Research Procedure:

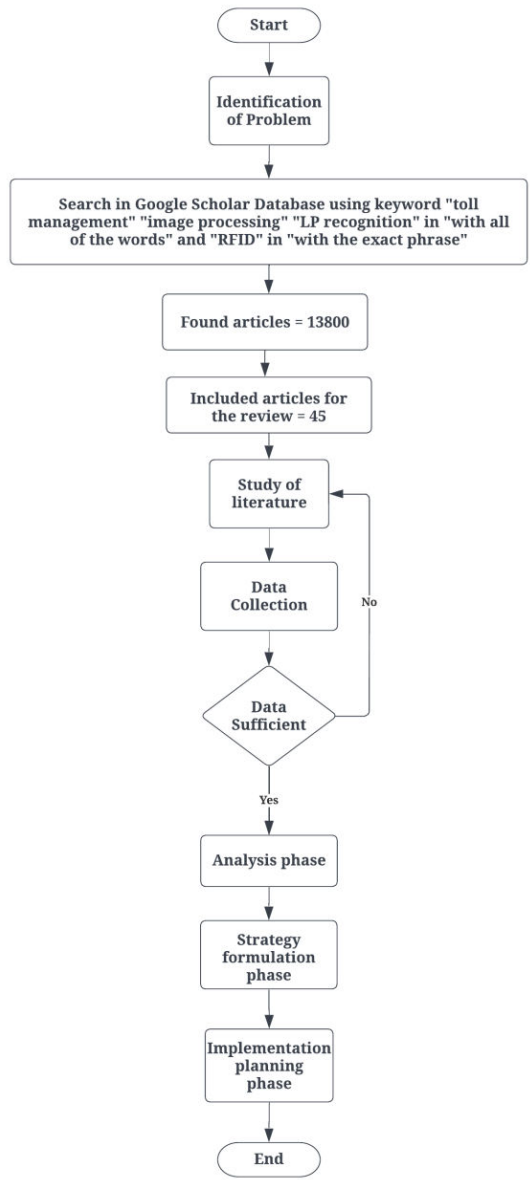

*Fig 1: Research Procedure Flowchart*

We first started by identifying the problem where the key issue of our project was not making vehicles wait while it passes the toll plaza. The primary problem we discovered in in the transport sector was transport making long queue lines and vehicles waiting for a long time. So, to mitigate that, we are implementing a system where through the help of image processing, we will click pictures of the number plate to capture RFID. It will save time as vehicles would not be waiting anymore thus traffic on road will reduce. In the next part of our research procedure, we used Google scholar to study different papers and databases using certain keywords. Keywords like "Automated toll management" in "with all the words" and "RFID" in "with the exact phrase" and "image processing", with another important keyword being "Image Processing" and lastly "license plate detection/recognition" including the above words. Then we filtered out around 13,800 articles and analyzed them thoroughly. After we completed noting and brainstorming, we went further to collect data. If the desired data we collected does not fulfill our requirements we will again go back to study the literature and if found sufficient we move to the analysis phase to formulate a strategy for our desired goal. Lastly, we move to our final stage where we implement the system. This research process had been followed throughout the project to reach our final goal of designing a system were number plate images get captured with help of image processing.

## 4.2 Overall System:

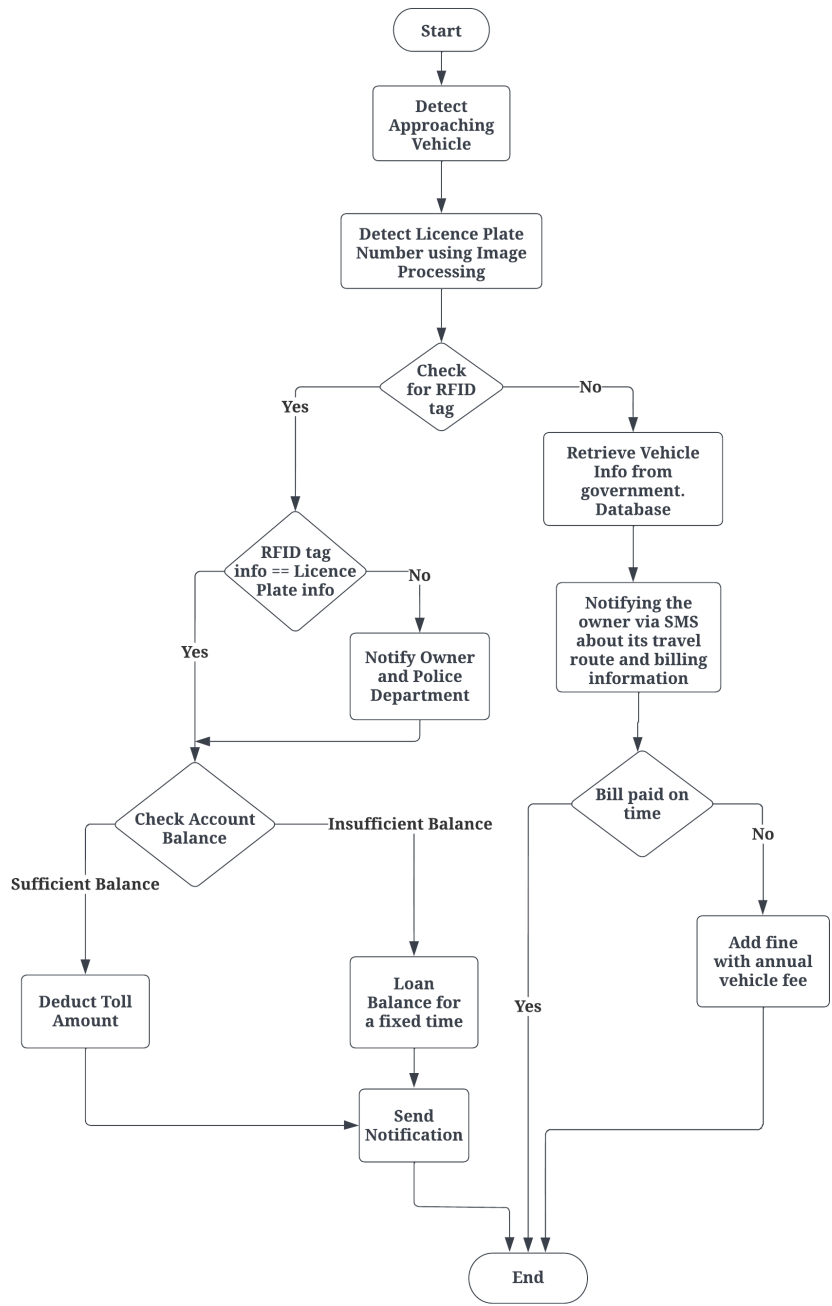

*Fig 2: Overall System Flowchart*

In this system whenever any vehicle approaches the toll plaza, when a car approaches the toll plaza, the approaching car is detected using a laser sensor. A RFID scanner will then search for an RFID tag. Then it will check if the vehicle's RFID (Radio-frequency identification) tag is activated or not. If the chip is activated then it will check the user's information from the database and deduct the balance from the car owner's account automatically. When the tag is inactive or the tag is not found a camera is triggered and the camera will a photo of the vehicle's license plate. Then according to the registered number plate, mail will be sent to the respective owner for paying the money within a certain amount of time to avoid a fine. Using this RFID tag, both time consumption is decreased and extra security measures are ensured. In this way, people don't have to stop at the toll gate even if they don't have registered for an RFID tag. Authority can still collect the fee from the unregistered users. This will completely remove any kind of traffic congestion at the toll gate. If a registered vehicle is stolen, the user can report the theft, and all toll plazas will be notified if that vehicle goes through. If the car goes through a toll plaza, local authorities will be notified of the vehicle's last known location, allowing them to monitor and locate it.

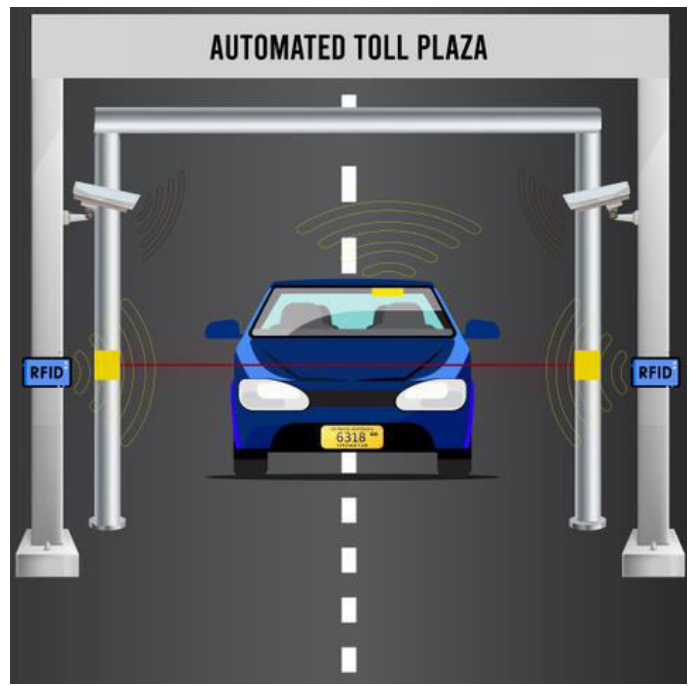

*Fig 3: Automated Toll Plaza*

There is software that can do these tasks only if the vehicle is registered and has an RFID tag on the windshield. If the vehicle is unregistered, the owner must manually pay the toll fee. But the difference between our system and the existing system is that our system will detect the vehicle owner even if the vehicle is unregistered to our system and send a mail to the respective owner informing him to pay the fee in time to avoid fines.

## 4.3 Image Processing:

Transfer Learning is a method in which a model developed for one task is adapted for another task that has been connected to it using the machine learning technique, using this method a pre-trained model is collected from open sources. Along with the Transfer Learning configuration, Pipelines were configured as well to detect objects. The pre-trained model has been trained with 411 images and 22 images for testing purposes. With the accomplishment of training, license plates were then detected. After successfully extracting the number plate Easy OCR is applied to properly distinguish numbers only, eliminating any other data on the license plate with OCR filtering. Moreover, after the images are saved and separated into a string format to a CSV file. In the end, a webcam or camera will be set to capture images conducting all the above procedures to extract the vehicle registration numbers from the number plate.

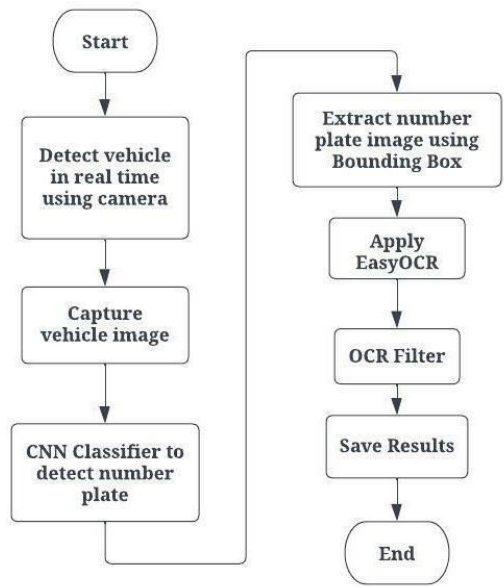

*Fig 4: Image Processing Flowchart*

## 4.4 Mobile Application Prototype

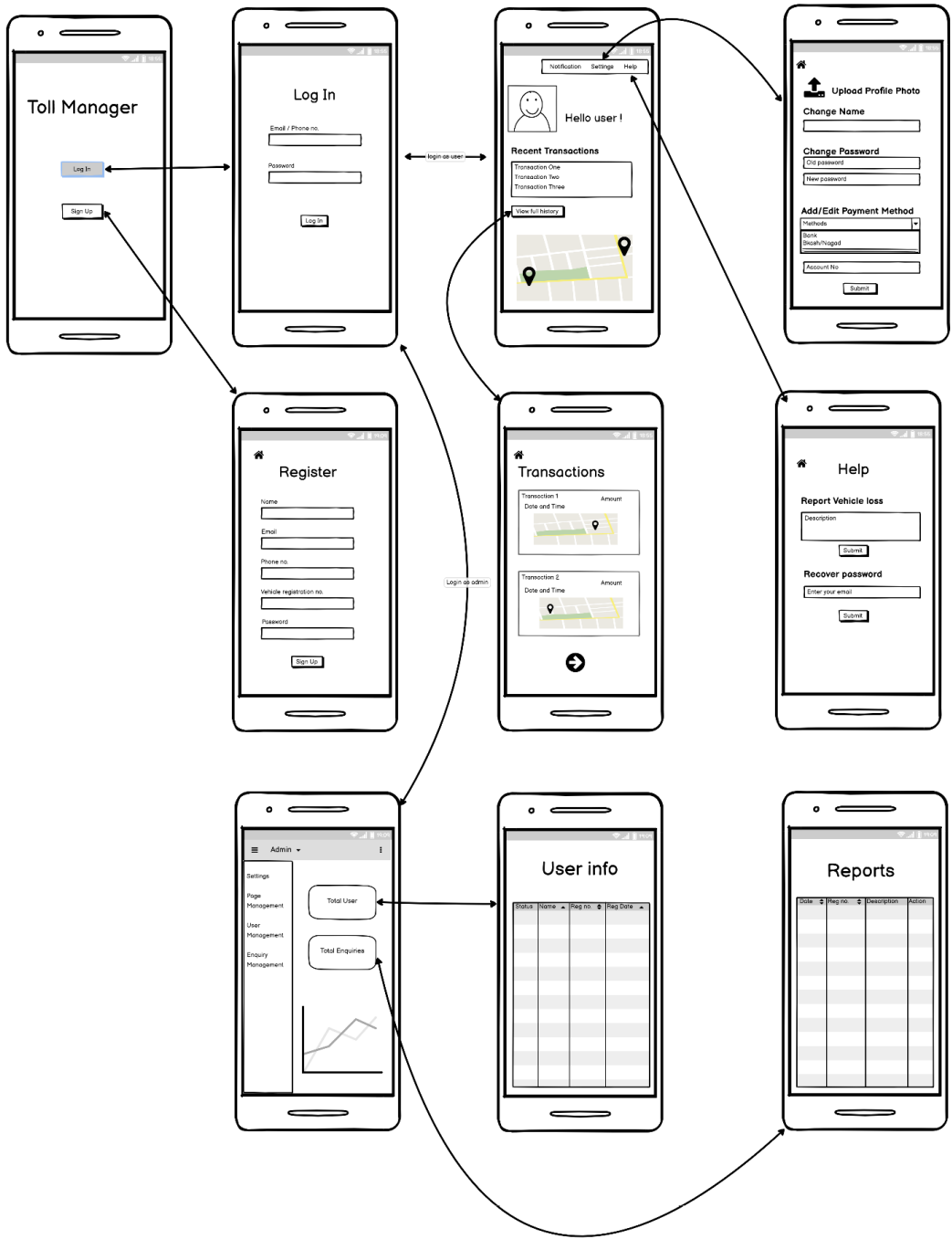

*Fig 5: Prototype Design*

### 4.4.1 Feature List

**Admin Features**

1. Sign In
2. Access to User Information
3. Change User Information
4. View total inquiries/reports
5. Respond to User Reports
6. Tracking Vehicles
7. Remove Users

**User Features**

1. Registration
2. Login
3. Checking Notification
4. Edit/Update Information
5. Change Password
6. Recover Password
7. Add Payment Method
8. Viewing Recent Transactions
9. Report Vehicle Loss

# 5. Results and Analysis:

# **Training and Evaluation Model**

Figure 1 depicts the learning rate and Figure 5-8 outlines the loss of training model.

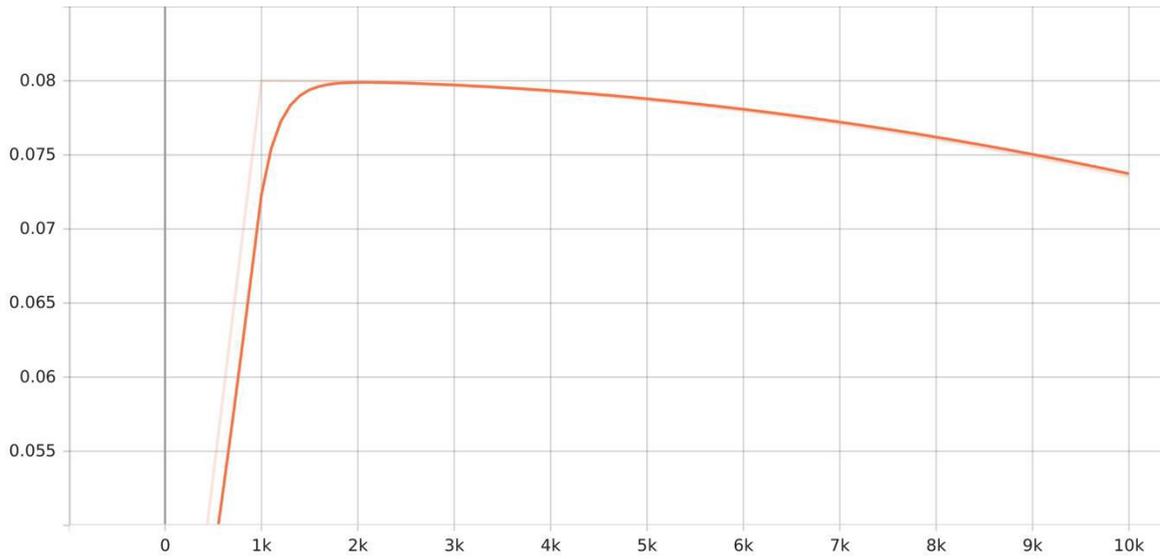
*Fig 6: Learning Rate*

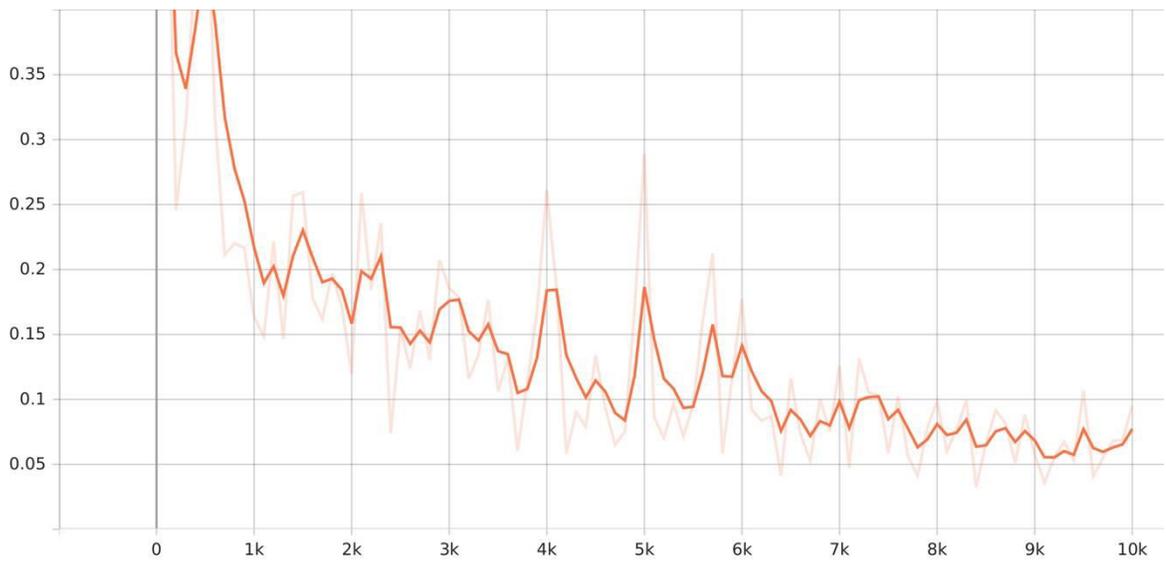
*Fig 7: Localization Loss*

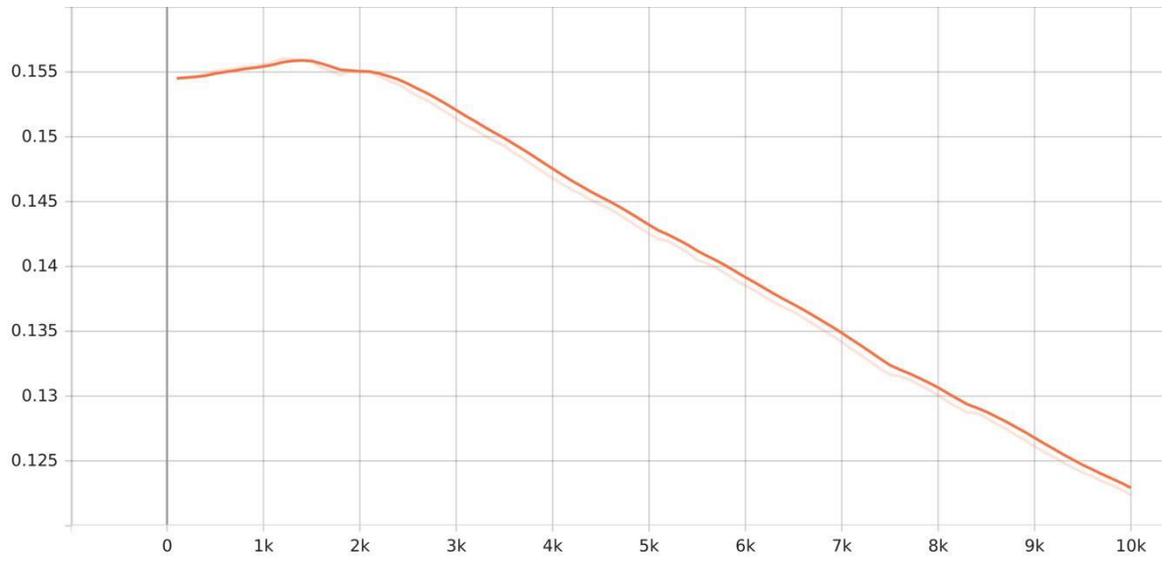
*Fig 8: Regularization Loss*

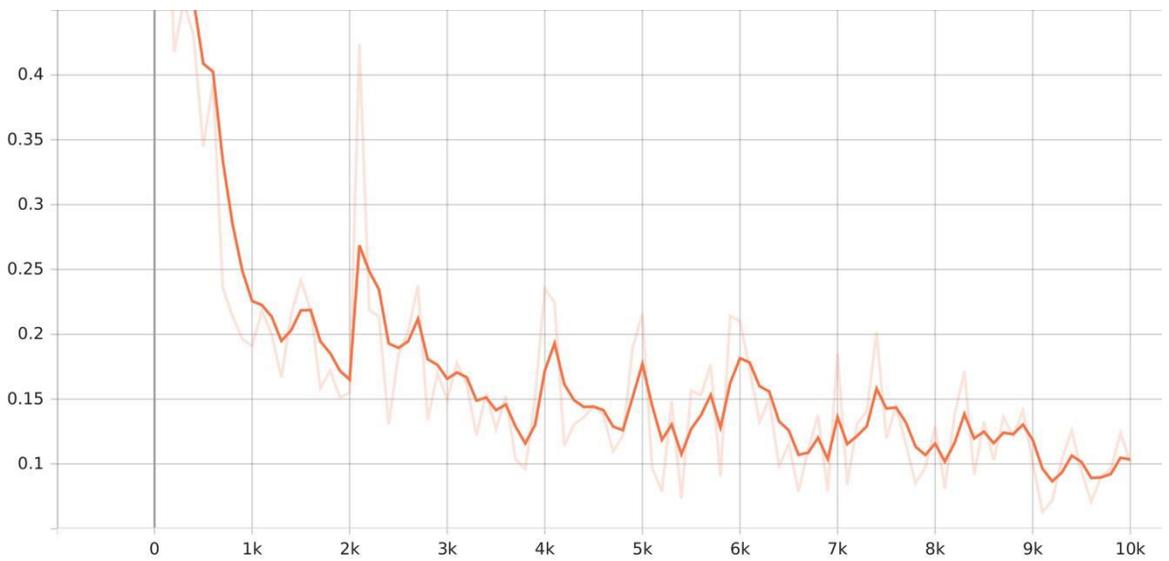
*Fig 9: Classification Loss*

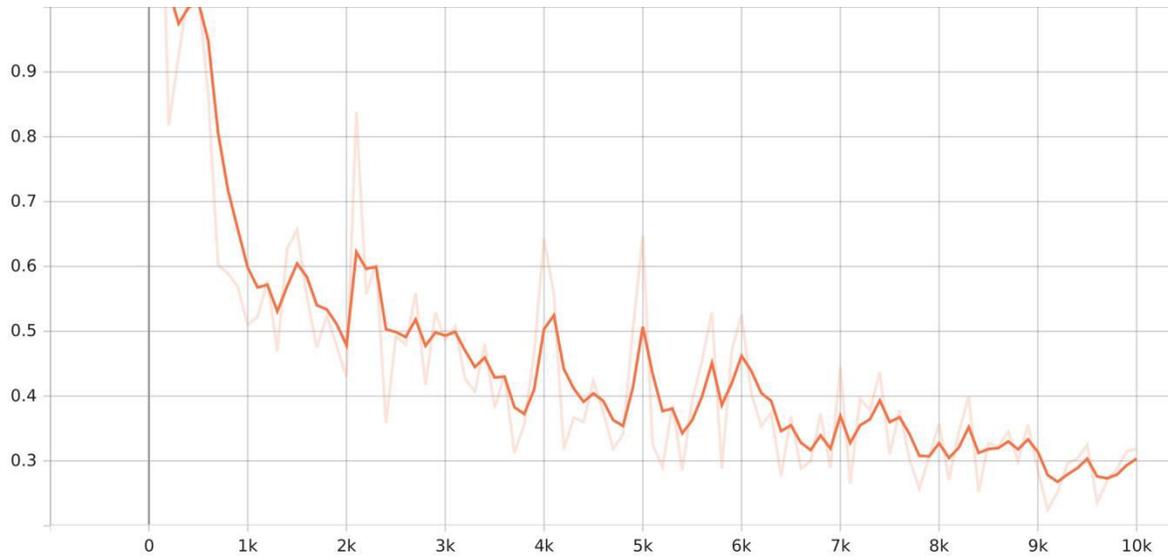
*Fig 10: Total Loss*

It can be perceived that within approximately 2000 steps the model reaches peak learning rate and total loss of the model also falls significantly. From 2000 steps onwards the learning rate decreases gradually and the same is the case for total loss till 10,000 steps. Table below summarizes all the data.

|  | After 10,000 Steps (Smoothed value) |
| --- | --- |
| Learning Rate | 0.07373 |
| Localization Loss | 0.0772 |
| Regularization Loss | 0.1229 |
| Classification Loss | 0.1035 |
| Total Loss | 0.3036 |

*Table 1: Smoothed values after 10,000 steps*

After evaluating the final model, it was determined that the mean average precision of the model is 0.529 and average recall of the model is 0.604 which are illustrated in Figure 9 and 10 respectively.

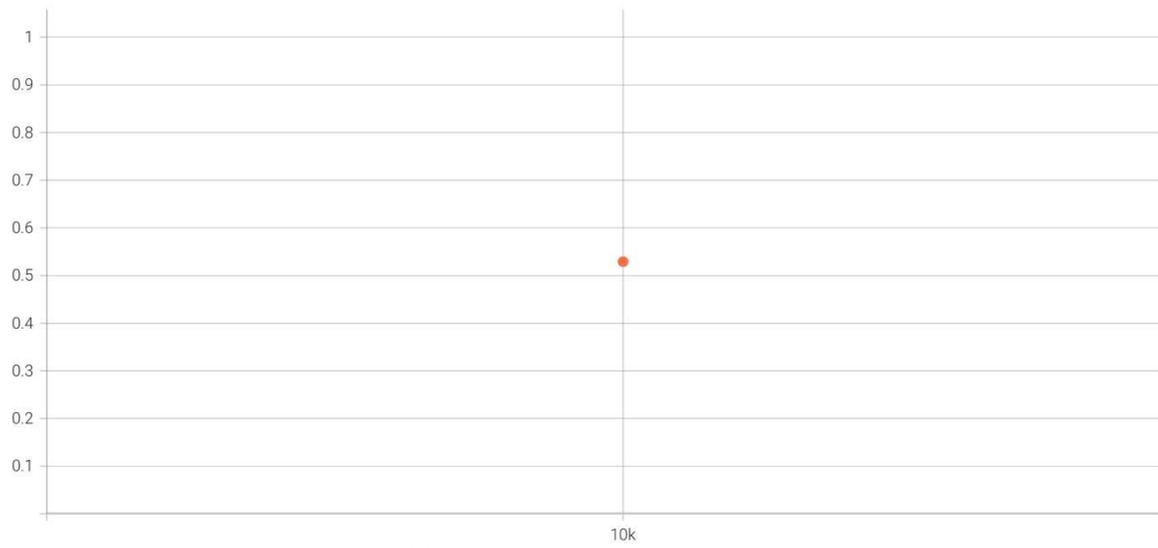

*Fig 11: Mean Average Precision*

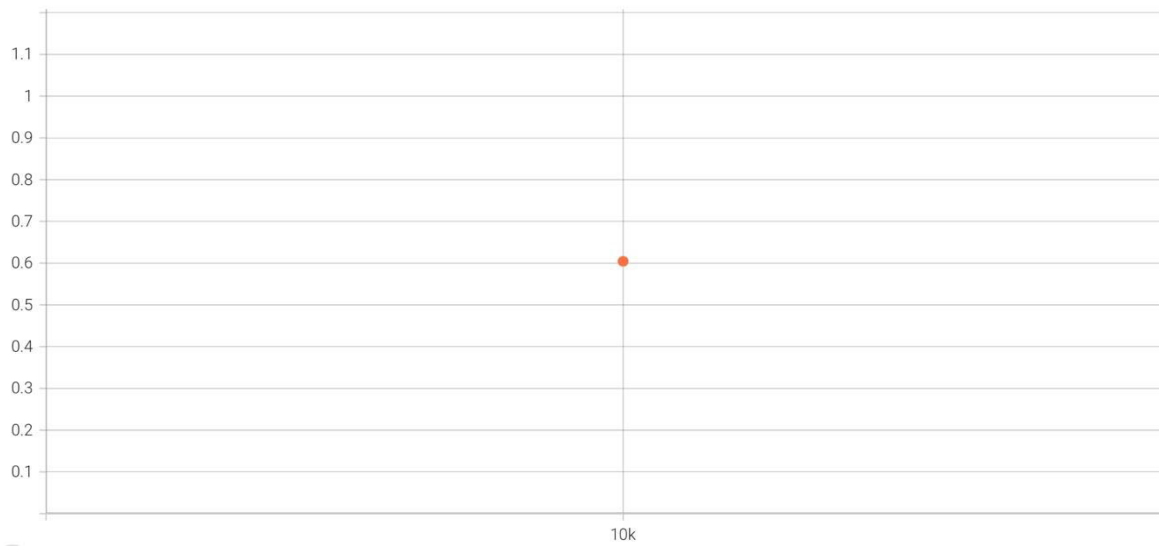

*Fig 12: Average Recall*

Table below gives detailed data about the mean average precision and average recall.

| Average Precision (area = all) | 0.529 |
|---|---|
| Average Precision (area = small) | 0.380 |
| Average Precision (area = medium) | 0.561 |
| Average Precision (area = large) | 0.651 |
| Average Recall (area = all) | 0.604 |
| Average Recall (area = small) | 0.475 |
| Average Recall (area = medium) | 0.653 |
| Average Recall (area = large) | 0.750 |

*Table 2: Mean average precision and average recall*

Below, in Figure 8 there are some test images showing the actual bounding box accuracy score and tested bounding box accuracy score of the model after evaluation.

**Model Accuracy Score**     **Actual Accuracy Score**

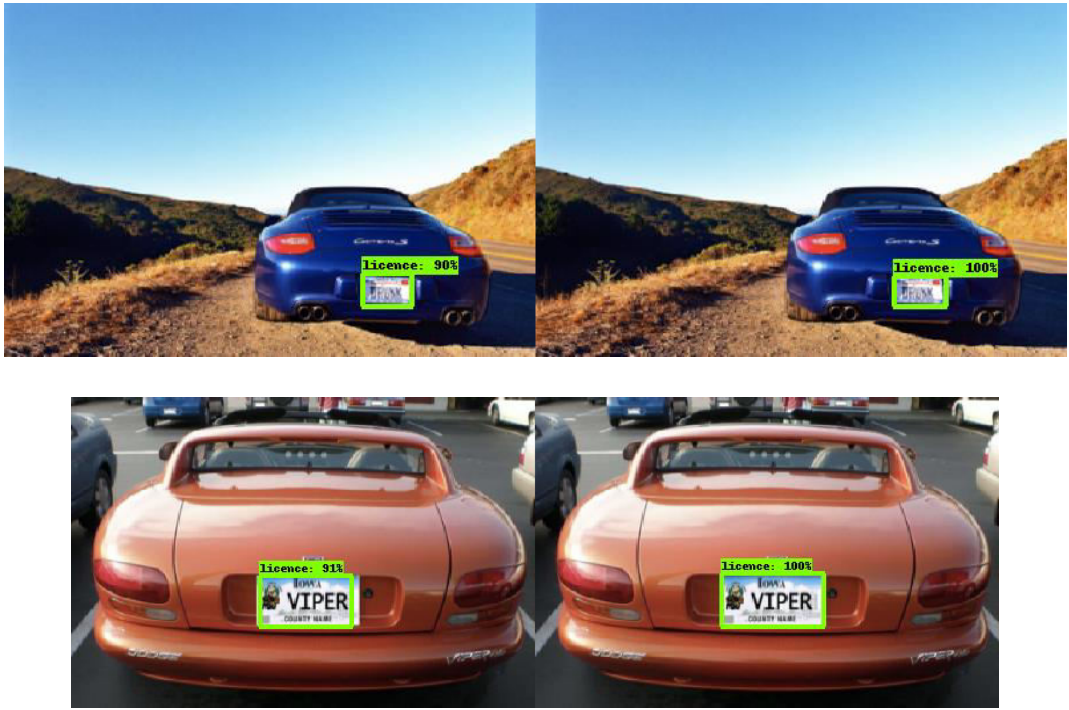

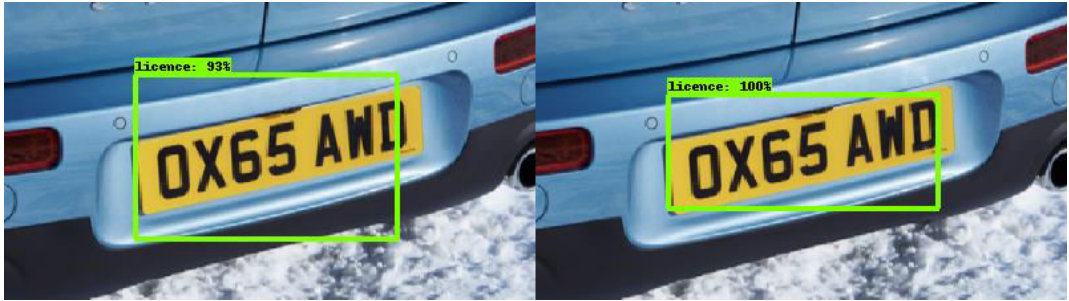

*Fig 13: Accuracy Evaluation*

It can be noticed that the trained model to detect license plate is performing with an accuracy of above 90% in these test images.

After detecting the license plate, its image is saved with a unique name and EasyOCR is applied on that image to extract the license plate number. This extracted number is saved in a csv file with corresponding image name. After that the license plate number retrieved using RFID tag and reader is matched with the extracted number of license plate from the camera to impose two factor authentications. This in turn increases accuracy of the overall system and helps detect any malicious vehicle passing through the toll plaza.

# 6. Conclusion and Future Work: -

The proposed system aims to ensure a fast and reliable automated toll management system that can also detect malicious vehicles, for example, any stolen vehicle or vehicle with a changed license plate or RFID tag. Moreover, when a vehicle passes through the toll plaza, images of number plates both front and back would be captured seamlessly without any halting of vehicles, hence avoiding queues and traffic jams on roads. On the other hand, payments are also being made digitally which is providing added security and lesser manpower would be needed as well. The model trained for license plate number recognition reaches a final total loss of 0.3036 after 10,000 steps. The model's mean average precision is 0.529 and average recall is 0.604 respectively. The project we have developed here is only a theoretical version not been physically implemented in real life. Due to a shortage of time, we were only able to capture images via webcam which could later be advanced to high-definition cameras providing clearer accuracy and precision. If we successfully implement our system some issues may arise along the way. One of them is a disruption in RFID tag detection and image capture which will eventually deplete the overall accuracy of the system. Secondly, machines that are set and integrated might also get stolen as no one will be present near the toll plaza to monitor or guard. Thirdly, at night time accuracy of images maybe not be as accurate as at daylight. Further study and improvement could be made in order to inculcate precision. Last but not least when our system will be incorporated with a large amount of data then how will it respond, it was not further tested. So, further study and evaluations are required to organize a bulk amount of data.